%% file: Template.tex
\documentclass{article}
\usepackage{spconf,amsmath,graphicx,hyperref}
\usepackage{booktabs}   
\usepackage{multirow}   
\usepackage{graphicx}   
\usepackage{array}      
\usepackage{microtype}
\ifx\paragraph\undefined
  \newcommand{\paragraph}[1]{\vspace{1em}\noindent\textbf{#1}}
\fi

\title{Forgive or Forget: Understanding the Context of Hate in Audio Retrieval Systems}
\name{ 
      Arghya Pal\textsuperscript{1}, Sailaja Rajanala\textsuperscript{1}, 
      Rapha\"{e}l C.-W. Phan\textsuperscript{1}, 
      Shekhar Nayak\textsuperscript{2}}
      
\address{\textsuperscript{1}School of Information Technology, Monash University, Malaysia campus\\
         \textsuperscript{2}University of Groningen, the Netherlands}
\begin{document}
%
\maketitle
\begin{abstract}
Handling toxic retrieval in text-to-audio systems is challenging due to contextual dependencies.  
Existing strategies (e.g., rephrasing, summarization) risk altering intent or omitting details.  
We propose a post hoc causal debiasing framework with a sentiment-controlled mediator to preserve semantic relevance while suppressing harmful speech.  
Our approach is model-agnostic and integrates seamlessly with existing retrieval pipelines.  
We introduce two variants: \emph{Forgive}, which re-ranks and filters toxic audio via logit adjustment, and \emph{Forget}, which generates counterfactual toxic prompts to mitigate harmful retrievals.  
Experiments show consistent toxicity reduction with minimal loss in retrieval accuracy, improving both safety and reliability.  
\end{abstract}
\begin{keywords}
Audio Retrieval, Hate Speech, Hate Audio, Causal Debiasing, Contextual Ranking, Toxic Speech
\end{keywords}
\input{sections/Introduction}
\input{sections/Methodology}
\input{sections/Results}

\bibliographystyle{IEEEbib}
\bibliography{refs}

\end{document}

%% file: sections/Introduction.tex
\section{Introduction}
\label{sec_intro}
Text-to-audio retrieval has gained increasing attention, driven by advances in large-scale audio encoders and transformer-based language models \cite{wu2023audio, oncescu2024dissecting, mei2024wavcaps}. Unlike traditional content-based retrieval systems that rely on manually crafted metadata \cite{elizalde2019cross, slaney2002semantic}, modern methods learn shared embedding spaces for text and audio. These approaches achieve strong performance on benchmarks such as AUDIOCAPS~\cite{kim-NAACL-HLT-2019} and CLOTHO~\cite{drossos2020clotho}, enabling applications in media search, video production, and accessibility.

Despite this progress, an underexplored challenge lies in the inadvertent retrieval of \emph{toxic} or \emph{hate} audio. Large-scale datasets inevitably include harmful speech (e.g., profanities, slurs, bullying) \cite{ali_2022_hate, maheshkumarnandwana_2024_voice}. While prior work has advanced text-based toxicity detection \cite{Davidson, costajussà2023mutox}, little attention has been devoted to toxicity mitigation in text-to-audio retrieval.

We propose a novel post-processing framework with two complementary strategies:
\begin{itemize}
    \item \textbf{Forget}: generates counterfactual toxic variants of queries and applies logit-averaging (akin to Noise2Noise \cite{lehtinen2018noise2noise}) to suppress consistently toxic retrievals, enforcing strict toxicity prevention at the model level.  
    \item \textbf{Forgive}: transcribes retrieved audio and applies toxicity classification, re-ranking outputs to filter harmful speech while retaining borderline but safe content.  
\end{itemize}

To evaluate effectiveness, we introduce three metric; \emph{Success Rate}, \emph{Accuracy}, and \emph{Sensitivity}—that jointly assess toxicity suppression and semantic relevance. Experiments on AUDIOCAPS and CLOTHO with three state-of-the-art models (ATNLL~\cite{wu2023audio}, TUAR~\cite{oncescu2024dissecting}, WavCaps~\cite{mei2024wavcaps}) show substantial improvements. The combined \emph{Forget+Forgive} approach consistently achieves superior non-toxic retrieval while preserving retrieval quality.

\begin{figure*}[!htbp]
    \centering
    \includegraphics[width=\textwidth]{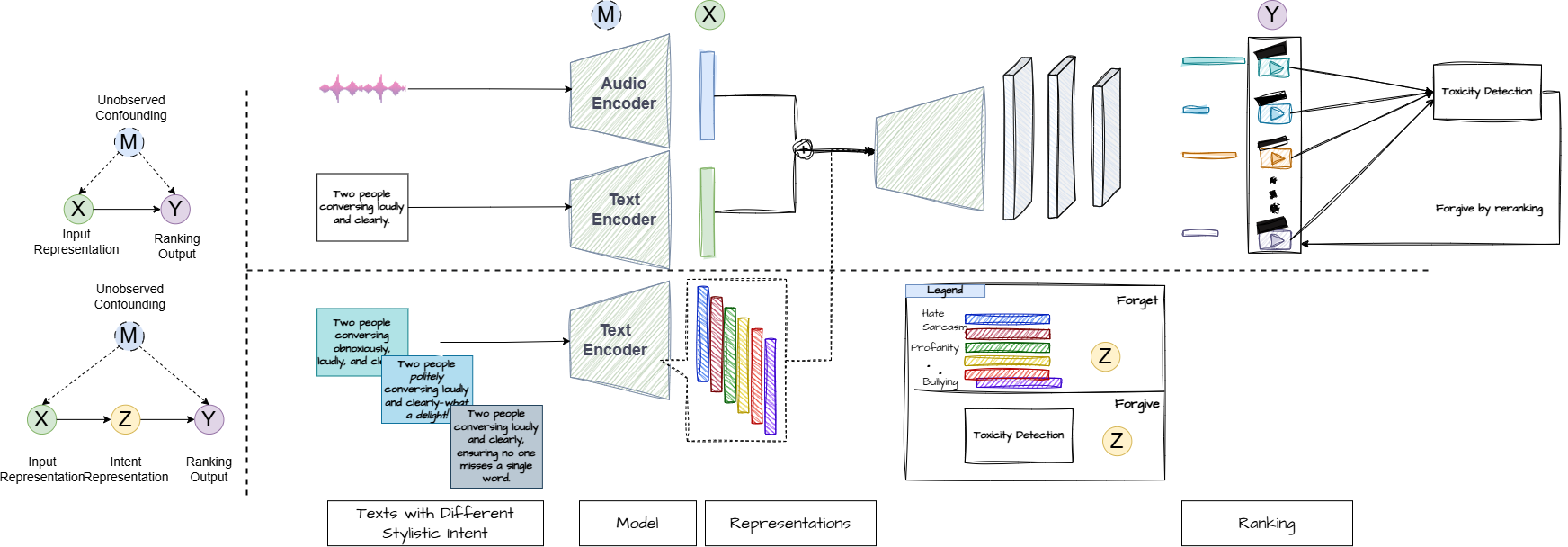} 
    \caption{\textbf{Overall methodology} The top part of the architecture shows the \textbf{Forgive} strategy where we embed the text using the text encoder of the respective models and the audio encoder captures audio semantics, with the Forgive strategy working by re-ranking toxic audios through softmax renormalization to lower their top-rank probability. At the bottom, we show the \textbf{Forget} pipeline where we visualize different counterfactual realizations of the input text prompt across six toxic categories, then apply Noise2Noise averaging of the logits to systematically suppress toxicity-driven rankings while preserving semantic relevance.}
    \label{fig:your_label}
    \vspace{-0.5cm}
    
\end{figure*}

%% file: sections/Methodology.tex
\section{Causal Debiasing Framework}
\label{sec:framework}

We introduce a post-hoc causal debiasing framework that mitigates toxic audio retrieval while preserving semantic relevance. The method applies two complementary \emph{front-door adjustment} mechanisms via a sentiment-controlled mediator, and can be integrated with any text-to-audio retrieval system.

\subsection{Problem Formulation and Causal Framework}

\noindent \textbf{Causal Structure.}  
In text-to-audio retrieval, the input prompt $T$ is embedded as $X$ by the model $M$, with ranking $Y$ determined jointly by $X$ and $M$: $X \leftarrow M \rightarrow Y$. Since $M$ is an unobserved confounder, we cannot directly estimate $P(Y|X)$ or apply standard backdoor adjustment~\cite{bellemare2024paper}.

\noindent \textbf{Front-Door Adjustment.}  
To address confounding, we introduce a mediator $Z$ representing stylistic variants of $X$. Specifically, $Z=\{Z_1,\ldots,Z_6\}$, where each $Z_i$ is semantically equivalent to $X$ but infused with one of six toxic intents: \emph{Hate, Profanities, Pornographic Language, Slurs, Violence, Bullying}. These variants are generated with an LLM while preserving the semantic core of the query. This yields a mediated path $X \rightarrow Z \rightarrow Y$, ensuring all causal effects from $X$ to $Y$ flow through $Z$ (see Fig.~\ref{fig:your_label}). By observing retrieval under toxic perturbations, we expose and suppress toxicity-driven biases. Using front-door adjustment:
\begin{equation}
P(Y | \text{do}(X)) = \sum_{z} P(Y | \text{do}(Z=z)) \, P(Z=z|\text{do}(X)),
\end{equation}
where $\text{do}(X)$ denotes interventions on prompts to produce toxic stylistic categories $Z$.

\subsection{Dual Mitigation Strategy}

\noindent \textbf{Forget: Counterfactual Toxic Prompt Generation.}  
The \emph{Forget} strategy makes the system statistically “forget” toxic content. Let $\ell_Y=f(X)$ denote retrieval logits for query $X$, and $\ell_Y^{(i)}=f(Z_i)$ for its toxic variants. We average logits across perturbations:
\begin{equation}
\ell_{f} = \frac{1}{|L|}\sum_{i}\ell_Y^{(i)},
\end{equation}
following the Noise2Noise principle~\cite{lehtinen2018noise2noise}. Since toxic biases fluctuate across variants but semantic relevance remains stable, averaging cancels toxic signals while retaining meaning. Thus, audio clips ranking highly only due to toxic phrasing are systematically suppressed.

\noindent \textbf{Forgive: Audio-Level Toxicity Filtering.}  
The \emph{Forgive} strategy applies lenient re-ranking at the audio level. Retrieved clips are first transcribed with Silero ASR~\cite{Veysov}, then classified with Detoxify~\cite{Detoxify}. Toxic audios are demoted via softmax renormalization over Top-$k$ candidates:
\begin{equation}
p(y_i) = \frac{\exp(\ell_{y_i})}{\sum_{j \in \text{Top-}k} \exp(\ell_{y_j})}.
\end{equation}
This retains borderline but safe content while penalizing explicitly toxic clips.

\noindent \textbf{Illustrative Example.}  
For the query \textit{``man arguing with woman''}, \emph{Forget} generates variants like \textit{``man insulting woman''} and averages logits, suppressing toxic-biased results. \emph{Forgive} then examines Top-10 retrievals, transcribes audio, and demotes those with hate speech, while preserving non-toxic arguments. Combined, \emph{Forget} prevents systematic toxic retrievals at the model level, and \emph{Forgive} ensures fine-grained safety at the audio level (see Fig.~\ref{fig:your_label}).

%% file: sections/Results.tex
\section{Experiments and Discussion}
\label{sec_results}

In this section, we present a series of experiments designed to evaluate our proposed postprocessing method for mitigating toxic audio retrieval in text-to-audio retrieval systems. 

\subsection{Datasets and Baseline Models}
\label{subsec:datasets}

We evaluate each model on three datasets, each featuring distinct characteristics and challenges:
\textbf{AUDIOCAPS}~\cite{kim-NAACL-HLT-2019}, \textbf{CLOTHO}~\cite{drossos2020clotho}. 
We use the designated \emph{test splits} of all three datasets in our experiments, ensuring a consistent evaluation across the different benchmark sets. We compare three state-of-the-art text-to-audio retrieval models: \textbf{ATNLL}~\cite{wu2023audio}, \textbf{TUAR}~\cite{oncescu2024dissecting}, \textbf{WavCaps}~\cite{mei2024wavcaps}. 

\subsection{Toxicity Analysis Matrices}
\label{subsec:toxicity}

Conventional retrieval metrics (e.g., precision, recall) offer limited insights into the toxicity of retrieved audio. To address this limitation, we propose three \emph{novel metrics} for evaluating non-toxic text-to-audio retrieval:

\begin{itemize}
    \item \textbf{Success Rate} measures the proportion of retrieved audio that is non-toxic:
    \[
    \text{Success Rate} = \frac{\left(\text{Top}@K \cap \text{Non-toxic}\right)}{K},
    \]
    where a higher Success Rate indicates more effective filtering of toxic content.
    
    \item \textbf{Accuracy (Acc)} quantifies how well retrieved non-toxic audio aligns with the \emph{target class}:
    \[
    \text{Acc} = \frac{\left(\text{Top}@K \cap \text{Non-toxic} \cap \text{class}\right)}{K \cap \text{class}},
    \]
    ensuring that the retrieval remains relevant while also being safe.

    \item \textbf{Sensitivity} evaluates the proportion of correctly retrieved non-toxic audio within the total occurrences of a given class:
    \[
    \text{Sensitivity} = \frac{\left(\text{Top}@K \cap \text{Non-toxic} \cap \text{class}\right)}{\#\text{class}},
    \]
    capturing the model’s ability to retrieve relevant non-toxic audio across the dataset.
\end{itemize}

These metrics provide a balanced view of how well a model filters toxicity \emph{and} maintains retrieval quality.

\subsection{Forget: Generating Toxic Prompts with an LLM}
\label{sec_forgive}
We use a LLM based counterfactual prompt generation technique to preserve semantic equivalence while injecting toxicity in input prompt. For instance, given a prompt:
\[
\textit{A man talking then a female talking when there is a train sound},
\]
we provide the LLM with an example Table~\ref{tab_toxic_methodology}, which details the prompt design and the one-shot prompting strategy used. To gauge the reliability of the LLM outputs, we manually inspected 40 randomly selected descriptions with their grounded sounds and the LLM’s toxicity assessment. We found an 85\% agreement between human evaluation and the LLM’s decisions, aligning well with prior observations in toxicity detection~\cite{costajussà2023mutox}. 
\begin{table}[htbp]
\centering
\resizebox{0.4\textwidth}{!}{%
\begin{tabular}{p{3.5cm} p{10.5cm}}
\hline
\textbf{Element} & \textbf{Content} \\
\hline
\textbf{Overall Task} &
\begin{minipage}[t]{\linewidth}
Generate synthetic text prompts that is semantically similar to input text prompts.
\end{minipage}
\\
\hline
\textbf{Input Prompt} &
\begin{minipage}[t]{\linewidth}
\textit{The input prompt is a regular english description using which we can retrieve audio files}
\end{minipage}
\\
\hline
\textbf{Evaluation Criteria} &
\begin{minipage}[t]{\linewidth}
1. \textbf{Correct} – The generated text prompt accurately reflects all relevant semantic information such as man, female, and train sound, in the correct order and timing. It adds toxicity such as Hate, Profanities, Pornographic Languages, Slurs, Violence, Bullying. \\
2. \textbf{Incomplete} – The generated text prompt partially captures semantic information but omits or slightly misrepresents toxicity details such as Hate,
Profanities, Pornographic Languages, Slurs, Violence, Bullying words. \\
3. \textbf{Wrong} – The generated text prompt is significantly inaccurate and doesnot entails Hate,
Profanities, Pornographic Languages, Slurs, Violence, Bullying., either describing non-toxic or grossly incorrect statements.
\end{minipage}
\\
\hline
\textbf{Guidelines for Correction} &
\begin{minipage}[t]{\linewidth}
- If necessary, revise the generated text description to maintain original vocabulary/style while fixing timing/sequence inaccuracies. \\
- Maintain or correct the chronological order of sounds according to the provided timestamps. \\
- Focus on whether the description’s timing aligns with the start/end points given for each sound. \\
- Use toxicity Hate,
Profanities, Pornographic Languages, Slurs, Violence, Bullying in the noun, pronoun, adjectives, and verbs.
\end{minipage}
\\
\hline
\textbf{Example (One-shot Prompting)} &
\begin{minipage}[t]{\linewidth}
\textbf{Input} \textit{Original audio description}: “A man talking then a female talking when there is a train sound” \\
\textit{Localized components and their start/end times}: \\
- revving: 2.154–10.02 \\
- a train sound: 0.0–10.02 \\[3pt]
\textbf{Output} \\
\textit{Evaluation}: Incomplete \\
\textit{Corrected description}: “A se**y man talking na**y then a sl**t female talking rubb**h when there is a Ch** Ch** train sound bo***”
\end{minipage}
\\
\hline
\textbf{Primary Considerations for LLM-Based Assessment} &
\begin{minipage}[t]{\linewidth}
- Ensure the LLM prioritizes the chronological order of sounds (start to end times) and toxicity addition. \\
- Check for missing sounds that occur within stated time intervals but are not described. \\
- Verify there are no extra sounds included that do not exist in the audio timeline. \\
- Preserve style: keep vocabulary/register consistent with the original description.
\end{minipage}
\\
\hline
\end{tabular}%
}
\caption{Methodology for generating text prompts that we described in Section 2}
\label{tab_toxic_methodology}

\end{table}

\subsection{Forgive: Retrieved Audio Understanding}
\label{subsec:forgive}

Our second strategy, \emph{Forgive}, operates on the \emph{retrieved} audio rather than the text prompts. After the candidate retrieval is performed by one of our baseline models (ATNLL, TUAR, or WavCaps), we run an audio processing pipeline to detect and filter toxic content. We use two key components: (1) Speech-to-Text Conversion: We leverage the Silero model~\cite{Veysov}, chosen for its lightweight architecture and speed, to convert the retrieved audio files into transcripts. To minimize transcription errors:
\begin{itemize}
    \item Audio files longer than 20 seconds are segmented to avoid errors in overly long contexts.
    \item Long silences and extraneous noise are removed.
    \item Output transcripts are normalized (e.g., consistent casing, punctuation).
\end{itemize}
and, (2) Toxicity Classification: once transcripts are obtained, we apply the Detoxify~\cite{Detoxify} classifier to identify toxic speech. Any audio file flagged as toxic within the \(\text{Top}@K\) candidates is removed, and we redistribute its retrieval score across the remaining non-toxic candidates. This score normalization ensures that the resulting ranking emphasizes non-toxic audio without sacrificing overall relevance.

\subsection{Quantitative Results}
\label{subsec:quant_results}

\begin{table*}[t!]
\centering
\begin{minipage}{0.48\textwidth}
\centering
\resizebox{\textwidth}{!}{%
\begin{tabular}{cl|ccc|ccc}
\hline
& & \multicolumn{3}{c|}{\textbf{Baseline}} & \multicolumn{3}{c}{\textbf{+ Both}} \\
\hline
Top@K & Model & SR & Acc & Sen & SR & Acc & Sen \\
\hline
\multirow{3}{*}{5}  
 & ATNLL~\cite{wu2023audio} & 0.35 & 0.42 & 0.45 & \textbf{0.90} & \textbf{0.58} & \textbf{0.80} \\
 & TUAR~\cite{oncescu2024dissecting} & 0.40 & 0.50 & 0.40 & \textbf{0.88} & \textbf{0.70} & \textbf{0.79} \\
 & WavCaps~\cite{mei2024wavcaps} & 0.55 & 0.60 & 0.50 & \textbf{0.89} & \textbf{0.82} & \textbf{0.85} \\
\hline
\multirow{3}{*}{10} 
 & ATNLL~\cite{wu2023audio} & 0.60 & 0.46 & 0.50 & \textbf{0.92} & \textbf{0.72} & \textbf{0.89} \\
 & TUAR~\cite{oncescu2024dissecting} & 0.45 & 0.55 & 0.43 & \textbf{0.92} & \textbf{0.74} & \textbf{0.80} \\
 & WavCaps~\cite{mei2024wavcaps} & 0.65 & 0.67 & 0.68 & \textbf{0.94} & \textbf{0.87} & \textbf{0.90} \\
\hline
\multirow{3}{*}{20} 
 & ATNLL~\cite{wu2023audio} & 0.64 & 0.50 & 0.54 & \textbf{0.97} & \textbf{0.87} & \textbf{0.95} \\
 & TUAR~\cite{oncescu2024dissecting} & 0.47 & 0.57 & 0.45 & \textbf{0.97} & \textbf{0.82} & \textbf{0.84} \\
 & WavCaps~\cite{mei2024wavcaps} & 0.68 & 0.69 & 0.71 & \textbf{0.97} & \textbf{0.89} & \textbf{0.93} \\
\hline
\end{tabular}
}
\caption{Performance on \textbf{AUDIOCAPS} dataset}
\label{tab:audiocaps_compact}
\end{minipage}
\hfill
\begin{minipage}{0.48\textwidth}
\centering
\resizebox{\textwidth}{!}{%
\begin{tabular}{cl|ccc|ccc}
\hline
& & \multicolumn{3}{c|}{\textbf{Baseline}} & \multicolumn{3}{c}{\textbf{+ Both}} \\
\hline
Top@K & Model & SR & Acc & Sen & SR & Acc & Sen \\
\hline
\multirow{3}{*}{5}  
 & ATNLL~\cite{wu2023audio} & 0.38 & 0.44 & 0.46 & \textbf{0.91} & \textbf{0.60} & \textbf{0.82} \\
 & TUAR~\cite{oncescu2024dissecting} & 0.41 & 0.52 & 0.41 & \textbf{0.89} & \textbf{0.72} & \textbf{0.80} \\
 & WavCaps~\cite{mei2024wavcaps} & 0.57 & 0.62 & 0.52 & \textbf{0.90} & \textbf{0.83} & \textbf{0.86} \\
\hline
\multirow{3}{*}{10} 
 & ATNLL~\cite{wu2023audio} & 0.60 & 0.49 & 0.51 & \textbf{0.93} & \textbf{0.75} & \textbf{0.91} \\
 & TUAR~\cite{oncescu2024dissecting} & 0.45 & 0.55 & 0.46 & \textbf{0.90} & \textbf{0.78} & \textbf{0.83} \\
 & WavCaps~\cite{mei2024wavcaps} & 0.64 & 0.68 & 0.69 & \textbf{0.95} & \textbf{0.88} & \textbf{0.91} \\
\hline
\multirow{3}{*}{15} 
 & ATNLL~\cite{wu2023audio} & 0.62 & 0.50 & 0.54 & \textbf{0.95} & \textbf{0.84} & \textbf{0.92} \\
 & TUAR~\cite{oncescu2024dissecting} & 0.46 & 0.58 & 0.47 & \textbf{0.92} & \textbf{0.80} & \textbf{0.84} \\
 & WavCaps~\cite{mei2024wavcaps} & 0.67 & 0.69 & 0.70 & \textbf{0.95} & \textbf{0.88} & \textbf{0.92} \\
\hline
\multirow{3}{*}{20} 
 & ATNLL~\cite{wu2023audio} & 0.68 & 0.52 & 0.58 & \textbf{0.97} & \textbf{0.80} & \textbf{0.94} \\
 & TUAR~\cite{oncescu2024dissecting} & 0.54 & 0.60 & 0.50 & \textbf{0.94} & \textbf{0.79} & \textbf{0.88} \\
 & WavCaps~\cite{mei2024wavcaps} & 0.72 & 0.71 & 0.74 & \textbf{0.98} & \textbf{0.91} & \textbf{0.95} \\
\hline
\end{tabular}
}
\caption{Performance on \textbf{CLOTHO} dataset}
\label{tab:clotho_compact}
\end{minipage}

\vspace{0.3cm}
\begin{minipage}{\textwidth}
\centering
\small
\textbf{Note:} SR: Success Rate, Acc: Accuracy, Sen: Sensitivity. Bold values show performance of our combined approach (Forget + Forgive). The combined method consistently outperforms baselines across both datasets and all Top@K values, with particularly strong improvements in Success Rate (toxicity reduction).
\end{minipage}
\vspace{-0.3cm}
\end{table*}
Tables 2 and 3 present a comprehensive evaluation of the three state-of-the-art models—ATNLL~\cite{wu2023audio}, TUAR~\cite{oncescu2024dissecting}, and WavCaps~\cite{mei2024wavcaps} across the AUDIOCAPS, and CLOTHO datasets, respectively. We use three non-toxic retrieval metrics: Success Rate $(\uparrow)$, Accuracy $(\uparrow)$, and Sensitivity $(\uparrow)$, at retrieval depths $\text{Top}@K = 5, 10, 15, 20$. For each model configuration, we show:
\begin{enumerate}
    \item Baseline model (no non-toxic mitigation),
    \item Logit adjustment for toxic audio suppression,
    \item Hate audio suppression using counterfactual prompts,
    \item Combined approach (logit adjustment + counterfactual prompts).
\end{enumerate}
All models tend to yield higher Success Rate with larger $K$. Baseline models exhibit the \emph{lowest} Success Rate, while strategies (ii)--(iv) gradually improve it, indicating more effective filtering of toxic content. Logit adjustment (ii) preserves relevance while reducing toxicity; counterfactual prompting (iii) is particularly effective at lower $K$, and the hybrid (iv) consistently yields the best overall Accuracy.

\subsection{Audio Distance Analysis}
\label{subsec:audio_distance}
Beyond toxicity, we also assess how our postprocessing approach impacts audio quality and consistency. We compute Spearman’s rank correlation coefficient (SC) to track \emph{degradation monotonicity} regarding audio intensity and quality, Pearson’s correlation coefficient (PC) to evaluate potential shifts in \emph{speech quality}, Non-Matching Audio Distance (NOMAD)~\cite{ragano2024nomad} to measure mismatches introduced by toxicity filtering. The audios after hate mitigation shows above 90\% similarity with ground truth on these scales.


\subsection{Conclusion}
\label{subsec:summary}
Handling sensitive topics such as hate and sarcasm in text-to-audio retrieval systems presents unique challenges due to their contextual
dependencies. We introduce a post hoc causal debiasing framework that applies front-door adjustment via a sentiment-controlled mediator to ensure that retrieved
audio is not biased towards harmful speech while maintaining semantic relevance. Our experiments leads to a marked reduction in toxic audio retrieval.

\subsection{Ablation Study}
\label{subsec:ablation}

We analyze the contributions of each component at $K{=}10$ (Table~\ref{tab:ablation}). \emph{Forget} achieves the strongest toxicity suppression (SR +0.30 to +0.40 over baseline), while \emph{Forgive} yields slightly better Accuracy and Sensitivity by preserving borderline but non-toxic content. Their combination delivers the best of both worlds.

\begin{table}[t]
\centering
\resizebox{\linewidth}{!}{%
\begin{tabular}{cl|ccc|ccc|ccc}
\toprule
& & \multicolumn{3}{c|}{\textbf{Baseline}} & \multicolumn{3}{c|}{\textbf{Forget}} & \multicolumn{3}{c}{\textbf{Forgive}} \\
\cmidrule(lr){3-5}\cmidrule(lr){6-8}\cmidrule(lr){9-11}
Top@$K$ & Model & SR & Acc & Sen & SR & Acc & Sen & SR & Acc & Sen \\
\midrule
\multicolumn{11}{c}{\textbf{AUDIOCAPS} ($K{=}10$)} \\
\midrule
 & ATNLL  & 0.60 & 0.46 & 0.50 & \textbf{0.88} & 0.66 & 0.80 & 0.86 & \textbf{0.70} & \textbf{0.83} \\
 & TUAR   & 0.45 & 0.55 & 0.43 & \textbf{0.85} & 0.72 & 0.77 & 0.84 & \textbf{0.74} & \textbf{0.79} \\
 & WavCaps& 0.65 & 0.67 & 0.68 & \textbf{0.91} & 0.83 & 0.87 & 0.89 & \textbf{0.85} & \textbf{0.88} \\
\midrule
\multicolumn{11}{c}{\textbf{CLOTHO} ($K{=}10$)} \\
\midrule
 & ATNLL  & 0.60 & 0.49 & 0.51 & \textbf{0.89} & 0.74 & 0.87 & 0.87 & \textbf{0.76} & \textbf{0.88} \\
 & TUAR   & 0.45 & 0.55 & 0.46 & \textbf{0.86} & 0.76 & 0.81 & 0.85 & \textbf{0.78} & \textbf{0.83} \\
 & WavCaps& 0.64 & 0.68 & 0.69 & \textbf{0.92} & 0.87 & 0.90 & 0.91 & \textbf{0.88} & \textbf{0.91} \\
\bottomrule
\end{tabular}}
\caption{Ablation at $K{=}10$. \emph{Forget} achieves stronger SR (toxicity suppression), while \emph{Forgive} maintains higher Acc/Sen.}
\label{tab:ablation}
\end{table}

\subsection{Robustness and Audio Quality}
We further assess robustness. (i) Human–LLM agreement on toxicity is 85\%, confirming reliable automatic generation. (ii) Spearman, Pearson, and NOMAD~\cite{ragano2024nomad} show $>$90\% similarity with references post-mitigation, indicating no perceptible degradation. (iii) Runtime overhead is modest: \emph{Forget} adds $\sim$6$\times$ retrieval passes (batched), and \emph{Forgive} requires lightweight ASR+classification only on Top@$K$ candidates.

\subsection{Summary}
Our framework substantially reduces toxic retrieval while preserving relevance. \emph{Forget} provides strict systemic suppression, \emph{Forgive} applies fine-grained filtering, and their combination achieves state-of-the-art safe text-to-audio retrieval.